%% file: main.tex
\documentclass{article}

\usepackage[dblblindworkshop,final,nonanonymous]{neurips_2026}

\workshoptitle{Physical World AI: Geometry, Characteristics, and Multimodal Sensing}

\usepackage[utf8]{inputenc}
\usepackage[T1]{fontenc}
\usepackage{url}
\usepackage{microtype}
\usepackage{graphicx}
\usepackage{booktabs}
\usepackage{amsmath}
\usepackage{amssymb}
\usepackage{array}
\usepackage{xcolor}
\usepackage{float}
\usepackage{needspace}
\usepackage{hyperref}

\hypersetup{
    hidelinks,
    pdfauthor={Arjun Subramanian},
    pdftitle={Counterfactual Action Evaluation, Observation Bottlenecks, and Representation Geometry in Joint-Embedding Predictive World Models}
}

\newcommand{\erank}{r_{\mathrm{eff}}}
\newcommand{\vnmse}{\mathrm{VN\mbox{-}MSE}}

\title{%
Counterfactual Action Evaluation, Observation Bottlenecks, and
Representation Geometry in Joint-Embedding Predictive World Models
}

\author{%
Arjun Subramanian\\
Computer Science and Artificial Intelligence Laboratory (CSAIL)\\
Massachusetts Institute of Technology\\
\texttt{arjun\_s@mit.edu}
}

\begin{document}
\maketitle
\begin{abstract}
Low latent prediction error does not establish that a world model distinguishes the consequences of its actions. We introduce an evaluation protocol that traces the same intervention through simulator state, raster observations, target embeddings, and predictor outputs. Exact simulator-state forks in a controlled deformable-physics testbed reveal distinct bottlenecks. Changed commands alter particle motion, yet \textbf{41.5\% of one-step raster pairs are identical}. Observation loss is not the whole explanation: among 579 high-visibility counterfactuals, median predictor-to-target response is 0.0051 and 0.0217 across two seeds, falling to 0.0027 and 0.0116 after variance normalization. An isotropic state perturbation matched to the target counterfactual embedding shift produces 190$\times$ and 53$\times$ larger predictor changes on the same visible pairs, isolating action-path under-use rather than a dead or globally shrunk predictor. MSE-only training gives 8.36$\times$ lower 10-step latent error in matched seeds, but in spectrally concentrated spaces; one VICReg target encoder is also strongly concentrated, so neither error nor rank alone certifies physical state. Finally, stiffness remains near chance even from full-resolution rasters and mechanical state while privileged material parameters decode perfectly, indicating weak identifiability under this excitation rather than encoder discard. These results motivate auditing physical effect, observation visibility, representation geometry, and action dependence separately.
\end{abstract}

\section{Introduction}
Latent dynamics support control from pixels~\citep{hafner2019planet}, and joint-embedding predictive architectures (JEPAs) learn representations through feature prediction~\citep{assran2023ijepa,bardes2024vjepa}. V-JEPA~2 extends this approach to action-conditioned prediction and planning~\citep{assran2025vjepa2}. For physical tasks, a useful representation must retain the geometry, motion, and material information needed to distinguish the consequences of different controls.

Several confounds complicate evaluation. Narrow representations can be easy to predict; connected action inputs can have little effect; decodable properties need not guide prediction. Crucially, rendering and quantization can hide a physically effective intervention. If observations do not distinguish alternative futures, weak sensitivity cannot be assigned entirely to the predictor.

\begin{figure}[t]
\centering
\includegraphics[width=\linewidth]{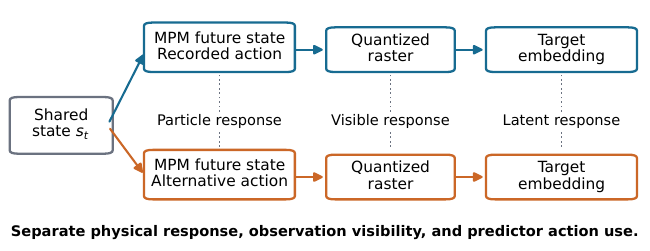}
\caption{\textbf{An action intervention must be measured at multiple levels.} From the same physical state, fork recorded and alternative commands. Particle differences need not survive rasterization or encoding. Separately compare the predictor outputs at fixed encoded state. Small latent sensitivity alone cannot locate the bottleneck.}
\label{fig:overview}
\end{figure}

We audit AC-JEPA in a two-dimensional deformable-object simulator. Cloning the full simulator state lets us trace counterfactual differences through particles, raster observations, target embeddings, and predictions (Fig.~\ref{fig:overview}). The contribution is an \emph{evaluation methodology and controlled case study} that makes observation visibility an explicit part of the diagnosis.

Three findings organize the paper. \textbf{Physical effects can disappear in the observations}: changed commands move particles while leaving many rasters identical. \textbf{Visible counterfactuals still reveal action-specific predictor attenuation}: response remains far below the target-embedding change after variance normalization, while matched state perturbations move the same predictor much more strongly. \textbf{Prediction error, spectral geometry, and property identifiability answer different questions}: low error can coexist with concentrated representations, and stiffness is not reliably identifiable even upstream of the encoder in this testbed. Each finding addresses a different requirement for physical grounding.

The study includes four matched VICReg action/no-action seed pairs, a two-seed objective comparison against MSE-only training, and geometry and stiffness probes. Conclusions are scoped to these checkpoints and this testbed; uncertainty and provenance details accompany the corresponding measurements.

\section{Related work}
\paragraph{Predictive representation learning.}
I-JEPA~\citep{assran2023ijepa} and V-JEPA~\citep{bardes2024vjepa} motivate learning representations through feature prediction; V-JEPA~2~\citep{assran2025vjepa2} motivates testing action-conditioned latent transitions. Non-contrastive methods address collapse through mechanisms such as bootstrapping~\citep{grill2020byol}, self-distillation~\citep{caron2021dino}, and variance/covariance regularization~\citep{bardes2022vicreg}. Our small model is not a reproduction of these large systems. We use a VICReg-style prediction objective and an invariance-only comparator to examine whether prediction accuracy, spectral diversity, and control dependence agree.

\paragraph{Physical world models.}
Graph Network-based Simulators learn particle dynamics~\citep{sanchezgonzalez2020gns}. PhysTwin reconstructs interactive deformable objects from videos~\citep{jiang2025phystwin}; PhysWorld combines physics-aware demonstration synthesis with a learned deformable-object world model~\citep{yang2025physworld}; DeformMaster integrates physical and neural modeling for interactive deformation~\citep{li2026deformmaster}. These works target physical reconstruction and simulation. Our complementary focus is how to audit physical claims in an observation-based latent model. Known simulator states and parameters allow controlled comparisons, without implying comparable realism or performance.

\paragraph{Evaluation scope.}
Rollout error, probes, and action ablations each answer a useful but different question. We add a paired physical reference for action replacement: the same intervention is measured before and after the observation operator. This distinguishes an ineffective physical contrast, an invisible observable contrast, and weak predictor response. The resulting protocol is relevant to benchmarks that expose geometry and dynamics through finite-resolution sensory representations.

\section{Testbed and completed training runs}
\label{sec:setup}
\paragraph{Simulator and observations.}
The data comprise 2,000 episodes with 16 observations each, giving 30,000 transitions. Initial shapes are disks or bars, and Young's moduli are $E\in\{18,70,220\}$ in simulator units. The NumPy simulator uses an MLS-MPM-style particle/grid update~\citep{hu2018mlsmpm}; inspection of its stress expression identifies a \emph{fixed-corotated} elastic form (Appendix~\ref{app:simulator}). We use that description rather than the older metadata's ``Neo-Hookean'' label.

The action is $a_t=(c_x,c_y,f_x,f_y)$: an actuator center and planar force command, resampled every three observations. Forces are distributed by a Gaussian normalized over particles; the center therefore redistributes the applied contribution spatially but does not set its total magnitude. A finite-difference audit at one-standard-deviation action scales finds that the force coordinates produce roughly twice the one-step position effect of the center coordinates, so we do not interpret all four components as equally effective. The actuator is smooth rather than a hard contact or grasp model. Each observation $x_t\in\mathbb{R}^{3\times64\times64}$ contains normalized density and two velocity channels. It is a physical raster, not camera RGB. Particle coordinates are rounded to raster centers, velocity is compressed by $\tanh$, and channels are quantized to uint8. The network receives neither material labels nor complete particle state.

\paragraph{Split and training.}
Episodes 0--1799 train the models; episodes 1800--1999 form the validation split used for the audits. Action z-score statistics use training episodes only. The corrected configuration disables random cropping ($\texttt{crop\_scale\_min}=1$); consistent photometric jitter and blur remain enabled during training. The corrected study includes VICReg with and without action conditioning for seeds 42--45, plus action-conditioned MSE-only for seeds 42/43. All ten checkpoints completed 200 epochs, and all reported comparisons use the final \texttt{last.pt}. Earlier cropped and smoke-scale runs are not pooled into these comparisons.

\paragraph{Model and objectives.}
The encoder is a four-block ViT with $8\times8$ patches, width 128, four attention heads, and a 128-dimensional CLS representation. The target encoder is updated by EMA with momentum 0.996. A four-block residual-MLP predictor concatenates the state with an embedded normalized action. The action-conditioned model has 874,368 trainable parameters, excluding its EMA copy. With stop-gradient denoted by $\operatorname{sg}$,
\begin{equation}
 z_t=f_\theta(x_t),\qquad z^-_{t+1}=\operatorname{sg}[f_\xi(x_{t+1})],\qquad
 \hat z_{t+1}=g_\psi(z_t,\tilde a_t).
 \label{eq:model}
\end{equation}
The VICReg-style objective is $25\mathcal L_{\mathrm{inv}}+25\mathcal L_{\mathrm{var}}+\mathcal L_{\mathrm{cov}}$, with prediction-target MSE, variance hinges, and off-diagonal covariance penalties~\citep{bardes2022vicreg}. Regularizers are evaluated on both branches, but only the prediction branch receives their gradients because the target is detached. MSE-only retains $\mathcal L_{\mathrm{inv}}$. Training uses AdamW, batch 128, learning rate $2\times10^{-4}$, 20 warmup epochs, cosine decay, weight decay 0.05, and gradient clipping at 1.0. Full settings appear in Appendix~\ref{app:repro}.

\paragraph{Evaluation data and provenance.}
The original observation dump is the canonical source for the final ten-checkpoint evaluation; all reported rollout errors use only episodes 1800--1999. A deterministic local replay of all 2,000 episodes reproduces the original dump exactly: 0 of 393,216,000 channel-pixel values differ and all recorded actions match. The original simulator-fork CSVs remain the source for physical counterfactual metrics. An earlier Linux recorded-action replay has a different observation checksum and is retained only as a cross-platform consistency check; it is not treated as byte-identical reconstruction. Appendix~\ref{app:provenance} records hashes and evaluator scope.

\section{Evaluation protocol}
\label{sec:protocol}
\subsection{Rollout error and representation geometry}
Starting from $z_0$, we recursively feed predicted latents and subsequent actions to $g_\psi$ for horizons $k=1,\ldots,10$. For $N=200$ validation episodes and dimension $d=128$, define
\begin{equation}
 \mathcal E_k=\frac{1}{Nd}\sum_e\|\hat z_{e,k}-z^-_{e,k}\|_2^2,\qquad
 \vnmse_k=\frac{\mathcal E_k}{d^{-1}\operatorname{tr}(\operatorname{Cov}_e[z^-_{e,k}])}.
 \label{eq:errors}
\end{equation}
The normalization uses population covariance at each horizon and is implemented explicitly in the new evaluator. It tests a global scale explanation, but it does not place independently learned representations in a common semantic space. Thus neither metric alone ranks physical fidelity across objectives.

For covariance eigenvalues $\lambda_i$ of validation latents, we report
\begin{equation}
 \erank=\frac{(\sum_i\lambda_i)^2}{\sum_i\lambda_i^2},\qquad
 q_1=\frac{\lambda_1}{\sum_i\lambda_i}.
 \label{eq:rank}
\end{equation}
We compute these quantities separately at horizons $k\in\{1,3,5,10\}$ for both prediction and EMA-target branches; pooled 1--10 values are supplemental only because pooling time-dependent distributions can inflate apparent diversity. The participation ratio measures spectral concentration rather than task usefulness. A high value can include irrelevant variation; a low value is not by itself proof of complete collapse.

\subsection{Action contrasts from the same physical state}
\label{sec:fork}
At each validation transition, snapshot particle positions, velocities, affine velocity fields, deformation gradients, and material parameters. The grid is reconstructed at every substep. Independent clones receive the recorded action, an action from another validation episode at the same time, or the training-split mean action (zero in normalized coordinates). Shuffles use a seeded permutation with fixed points repaired to ensure every source episode changes. The alternative command is held for the remainder of its three-observation block, then the recorded sequence resumes. Horizons $h=1,3,5$ contain 3,000, 2,600, and 2,200 valid transition/horizon pairs, respectively.

We measure RMS particle position and velocity differences, normalized raster MSE, and the fraction of exactly identical rasters. With $R$ the observation operator and $F_h$ the simulator rollout, the target and predictor contrasts are
\begin{align}
 D_{\mathrm{tgt},h}&=d^{-1}\|f_\xi(R(F_h(s_t,a))) - f_\xi(R(F_h(s_t,a')))\|_2^2,\label{eq:targetcontrast}\\
 D_{\mathrm{pred}}&=d^{-1}\|g_\psi(z_t,\tilde a_t)-g_\psi(z_t,\tilde a'_t)\|_2^2.\label{eq:predcontrast}
\end{align}
The compact notation $F_h(s_t,a)$ includes the command-hold and resume rule above. Predictor contrast is one-step; it is not presented as a multi-step counterfactual accuracy metric.

We additionally report the change in one-step error against the \emph{recorded} future,
\begin{equation}
 \Delta_{\mathrm{shuffle}}=100\frac{\mathcal E_{\mathrm{shuffled}}-\mathcal E_{\mathrm{recorded}}}{\mathcal E_{\mathrm{recorded}}}.
 \label{eq:shuffle}
\end{equation}
Both predictions use the same encoded current state. We report all transitions and command-change times $t\in\{3,6,9,12\}$ separately. The latter prevent a newly sampled command from already being reflected in the pre-action velocity. A trained no-action model is a distinct optimization ablation; it is not a substitute for changing inputs within a checkpoint.

To test whether raw response attenuation is merely a scale effect, we also normalize each contrast by the corresponding mean latent variance over validation transitions, $\bar D_{\mathrm{pred}}=D_{\mathrm{pred}}/\bar v_{\mathrm{pred}}$ and $\bar D_{\mathrm{tgt}}=D_{\mathrm{tgt}}/\bar v_{\mathrm{tgt}}$. As a matched state-sensitivity control, we perturb $z_t$ in a fixed isotropic random direction with $\ell_2$ norm $\sqrt{dD_{\mathrm{tgt}}}$ and measure $D_{\mathrm{state}}$, the resulting predictor-output MSE at fixed action. The response ratio $D_{\mathrm{pred}}/D_{\mathrm{tgt},1}$ is reported only where the denominator exceeds $10^{-8}$, with exclusions disclosed. All ratios compare response magnitudes, not response directions or fractions of correctly captured physics; medians are used because small denominators make means unstable.

\subsection{Property probes and uncertainty}
Frozen linear probes predict geometry, stiffness class, and contact quadrant from each rollout horizon. They train on 1,800 episodes for 80 full-batch AdamW steps at learning rate $10^{-2}$ and are evaluated on 200 validation episodes. Probe initialization is fixed; horizon accuracies are summarized descriptively. Uniform-guess rates are 50\%, 33.3\%, and 25\%, but contact classes are imbalanced. A constant classifier selected from the training majority at each horizon achieves \textbf{40.1\%} mean validation contact accuracy; this is the relevant simple reference.

For physical identifiability, we supplement the latent probes with upstream positive controls. Fixed linear and one-hidden-layer MLP readouts are applied to mechanical state $(x,v,C,F)$, full-resolution float rasters before uint8 casting, full-resolution uint8 rasters, and their histories. Geometry and shuffled-label controls test probe sanity; privileged $(\mu,\lambda)$ parameters test whether the readout can recover the material class when the defining variables are supplied. These controls use fixed settings without validation-based architecture selection.

Counterfactual uncertainty is estimated by resampling whole episodes, conditional on each checkpoint. The action-distance correlation uses the same episode-level bootstrap. Two-seed objective comparisons display both seeds directly rather than presenting a half-range as an uncertainty interval. Appendix~\ref{app:secondary} specifies the bootstrap and conditional-ratio calculations.

\section{Results}
\label{sec:results}
\subsection{Physical action effects are often hidden by the observation operator}
In the original fork experiment, one-step shuffled commands produce median RMS position change $1.19\times10^{-5}$ and velocity change $2.39\times10^{-3}$ in simulator units. Nevertheless, 41.53\% of the resulting raster pairs are exactly identical. These 1,246 pairs have nonzero particle differences but identical uint8 observations. Mean-action replacement yields 57.93\% identical pairs. Median shuffled position change is only $7.50\times10^{-4}$ pixels at the raster's 63-pixel coordinate span, so the physical intervention is real but very small. Across all shuffled pairs, 48.5\% preserve the density channel while changing velocity channels; only 9.9\% change density.

\begin{figure}[t]
\centering
\includegraphics[width=\linewidth]{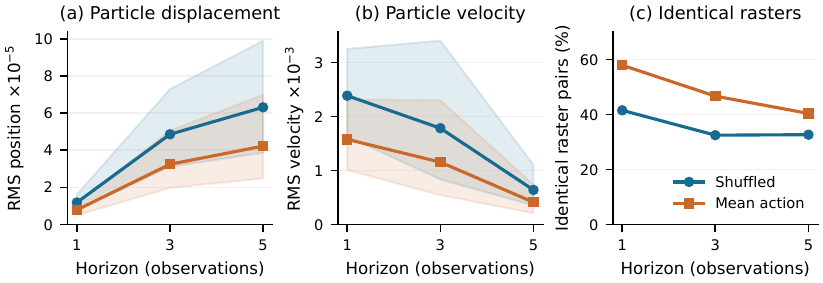}
\caption{\textbf{Physical response and observable response differ.} Original simulator forks across 200 validation episodes. Panels (a,b) show medians with interquartile bands, not confidence intervals. Panel (c) includes exact raster equality. Horizons 1/3/5 use 3,000/2,600/2,200 pairs; they do not have identical temporal support. Position differences grow, while many rasters remain unchanged.}
\label{fig:physical}
\end{figure}

At horizons 3 and 5, median shuffled position differences increase to $4.86\times10^{-5}$ and $6.31\times10^{-5}$, but identical-raster fractions remain 32.50\% and 32.68\% (Fig.~\ref{fig:physical}). The actuation and damping schedule therefore produces real but often subpixel differences over the short sequence. The $0.92$ per-substep velocity multiplier gives a nominal $0.92^8=0.513$ factor per observation before other dynamics; an impulse control retains only 0.163 of its horizon-1 RMS velocity by horizon 5. All 3,000 baseline transitions satisfy the recorded proximity rule. Because the actuator normalizes Gaussian weights, this rule is not a binary contact/no-contact mechanism and supplies no useful stratification here.

This result limits the interpretation of a naive action shuffle. An action-conditioned model cannot be expected to distinguish two \emph{identical} target observations through a reconstruction-style target metric. The physical reference reveals an observation bottleneck that would otherwise be misattributed entirely to model failure.

\subsection{Visible counterfactuals isolate action-path under-use}
\label{sec:sensitivity}
\begin{figure}[t]
\centering
\includegraphics[width=\linewidth]{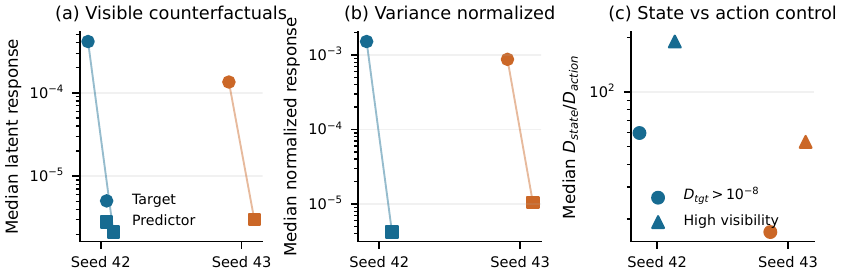}
\caption{\textbf{Visible counterfactuals separate observation loss from action-path under-use.} Original simulator forks, 579 high-visibility transitions per seed. (a) Median target and predictor response on the top third of nonzero raster-MSE contrasts. (b) Dividing by each branch's mean latent variance makes the gap larger, not smaller. (c) Perturbing the encoded state in an isotropic direction with $\ell_2$ norm matched to the target counterfactual shift produces much larger predictor-output changes than replacing the action. Ratios compare magnitudes, not response direction or control quality.}
\label{fig:visiblecontrol}
\end{figure}

The predictor responds weakly even after conditioning on interventions that survive the observation operator. Define the high-visibility subset as the top third of nonzero one-step raster MSE values ($n=579$). For seed 42, median target contrast is $4.13\times10^{-4}$ while predictor contrast is $2.14\times10^{-6}$; for seed 43 they are $1.35\times10^{-4}$ and $3.04\times10^{-6}$. The corresponding median $D_{\mathrm{pred}}/D_{\mathrm{tgt}}$ ratios are 0.0051 and 0.0217 (Fig.~\ref{fig:visiblecontrol}a).

Global predictor shrinkage does not explain this gap. Predictor variance is larger than target variance (0.510 vs. 0.272 for seed 42; 0.289 vs. 0.154 for seed 43), and variance normalization reduces the high-visibility median ratios further to 0.0027 and 0.0116 (Fig.~\ref{fig:visiblecontrol}b). A matched state perturbation gives a complementary control: among pairs with $D_{\mathrm{tgt}}>10^{-8}$, median $D_{\mathrm{state}}/D_{\mathrm{action}}$ is 59.5$\times$ and 16.9$\times$ for seeds 42/43; on the high-visibility subset it rises to 190$\times$ and 53$\times$, with state response larger than action response on 97.2\% and 96.7\% of pairs. Thus the predictor is not globally inert; these checkpoints are specifically much less sensitive to supplied action than to comparably scaled state changes.

Action-distance analyses support the same diagnosis without defining the primary claim. Normalized action distance correlates with RMS physical displacement (Spearman $\rho=0.684$; 95\% episode-bootstrap interval $[0.659,0.707]$). In the largest distance quintile, median target contrast is $9.03\times10^{-5}$ versus predictor contrast $5.17\times10^{-6}$ for seed 42; seed 43 gives $3.98\times10^{-5}$ versus $7.01\times10^{-6}$. Correct response direction and downstream control remain untested.

\subsection{Four seeds show no consistent action-conditioned rollout advantage}
\begin{figure}[t]
\centering
\includegraphics[width=\linewidth]{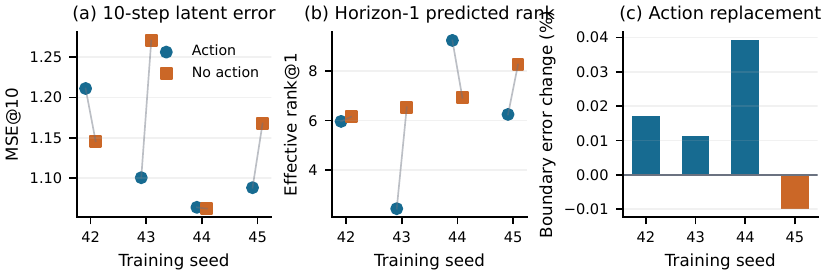}
\caption{\textbf{Four-seed action/no-action comparison on the original validation dump.} Each point is a final epoch-200 checkpoint; gray segments connect the matched action/no-action models within a seed, not across seeds. (a) MSE@10 ordering changes with seed. (b) Horizon-1 predicted rank also varies. (c) Replacing actions at command-change times produces only small signed changes in recorded-future error.}
\label{fig:seeds}
\end{figure}

Across the four matched VICReg seeds, action/no-action MSE@10 ordering flips: no-action is lower for seeds 42 and 44, while action-conditioned is lower for seeds 43 and 45 (Fig.~\ref{fig:seeds}a). Horizon-1 predicted rank also varies strongly across seeds (Fig.~\ref{fig:seeds}b). We therefore detect no consistent action-conditioned rollout-error or spectral advantage at $n=4$.

At command-change boundaries, replacing the supplied action changes recorded-future error by only $+0.017\%$, $+0.011\%$, $+0.039\%$, and $-0.010\%$ for seeds 42--45. The sign flip and seed variation make this a null architectural comparison rather than evidence that action conditioning improves forecasting. No-action predictions have zero supplied-action contrast by construction.

\subsection{Low error and spectral concentration must be audited separately}
\begin{figure}[t]
\centering
\includegraphics[width=\linewidth]{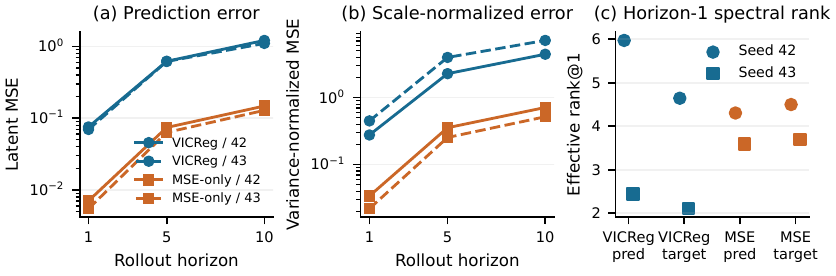}
\caption{\textbf{Matched objective comparison, seeds 42/43, on the original validation dump.} (a,b) Individual seed curves are shown directly; no two-seed error bar is interpreted as sampling uncertainty. MSE-only has much lower latent error with and without target-variance normalization. (c) Horizon-1 effective rank is concentrated in the MSE runs, but VICReg is not immune: seed 43 has a target rank near 2.1. Neither low error nor spectral breadth alone certifies useful physical state.}
\label{fig:objectives}
\end{figure}

Matched MSE-only runs achieve much lower latent rollout error: at horizon 10 their mean is 0.1382 versus 1.1558 for VICReg, an 8.36-fold difference, and the ordering survives target-variance normalization (Fig.~\ref{fig:objectives}a,b). But the learned spaces are strongly concentrated. At horizon 1, MSE predicted ranks are 4.30 and 3.59 out of 128 dimensions; their target ranks are 4.50 and 3.69. VICReg is not uniformly broader: predicted ranks are 5.97 and 2.44, while target ranks are 4.64 and 2.11 for seeds 42/43 (Fig.~\ref{fig:objectives}c). Pooled 1--10 rank can substantially overstate single-horizon diversity (e.g., seed-42 VICReg prediction: 12.27 pooled vs. 5.97 at horizon 1). Thus the 8.36$\times$ error gap is a metric disagreement, not evidence that MSE-only forecasts physics better; spectral diagnostics themselves require horizon- and branch-specific inspection.

\subsection{Geometry is a positive control; stiffness is not identifiable under this excitation}
Across four corrected VICReg action seeds, predicted-latent geometry accuracy is 99.75--100\%, an expected positive control for disk-versus-bar shape in the density channel. Contact accuracy is 38.35--39.70\%, below the 40.1\% training-majority reference, so it does not establish useful control information.

The new upstream controls change the interpretation of stiffness. Fixed readouts of full-resolution float and uint8 rasters remain at 33.0--33.5\% across tested frames and histories, essentially the 33.5\% validation-majority baseline. Mechanical state $(x,v,C,F)$ is also unstable, ranging from 28.5\% to 39.5\% across readouts/times while geometry remains 100\%. In contrast, an MLP given privileged $(\mu,\lambda)$ material parameters recovers stiffness at 100\%. The failure therefore occurs upstream of the JEPA encoder: under this actuation and short horizon, stiffness is not reliably identifiable from the observed mechanical trajectory or raster. We do not interpret near-chance latent probes as evidence that the encoder discarded an otherwise observable material signal. Appendix~\ref{app:probes} reports the full hierarchy.

\section{Discussion and limitations}
\label{sec:limitations}
The central lesson is to validate and trace the \emph{contrast} before diagnosing the model. Here the environment attenuates action effects strongly, the observation operator erases many remaining differences, and the predictor adds a distinct action-specific attenuation on counterfactuals that are already visible. Variance normalization and the matched state perturbation rule out a simple dead-predictor or global-scale explanation, but they do not test whether the predicted response points in the physically correct direction. Directional alignment between predicted and target counterfactual contrast vectors, or ranking the recorded future against alternative futures, would provide complementary tests of action-conditioned prediction quality.

The testbed is one short-horizon 2D setting. Its normalized actuator delivers nearly fixed total impulse while redistributing it by center, force coordinates are about twice as influential as center coordinates at typical scale, and $0.92$ substep damping rapidly removes velocity. Density/velocity rasterization, quantization, held commands, and photometric augmentation can further favor state-based prediction. The actuator does not model hard contact. Only two checkpoints support exact simulator-fork latent analyses and the matched objective comparison, despite four-seed architectural coverage. Episode-bootstrap intervals omit training-seed uncertainty. Validation episodes are reused for diagnostics; there is no untouched final test split or formal equivalence test. Retraining under stronger or more persistent action effects would help determine how much of the observed attenuation is specific to the present excitation and observation regime.

The original observation dump is locally bitwise reproducible, but an earlier cross-platform Linux replay has a different checksum and remains separately identified. The stiffness controls show weak identifiability under this excitation rather than encoder-specific information loss, while the privileged-material probe also demonstrates that probe capacity matters. The study includes no planning benchmark, decoded physical-state forecast comparison, real-video experiment, or sim-to-real validation. In particular, downstream closed-loop control remains untested, so the present diagnostics do not establish whether stronger or better-aligned action responses would improve decision making. Neither spectral breadth nor nonzero action sensitivity establishes physical understanding.

\section{Conclusion}
The simulator-fork audit localizes three distinct failure modes: small physical effects can be erased by rasterization, predictor action response can remain strongly attenuated even for visible counterfactuals, and latent-error/spectral metrics can disagree without identifying useful physical state. Upstream stiffness controls further show why property failures should be checked for identifiability before being attributed to representation loss. Physical-world evaluation should therefore trace interventions through physical state, observations, representations, and action-conditioned predictions rather than rely on latent rollout error alone.

\clearpage
\bibliographystyle{plainnat}
\bibliography{references}
\clearpage
\appendix
\input{appendix.tex}
\end{document}

%% file: appendix.tex
\section{Simulator and observation operator}
\label{app:simulator}
The simulator maintains $(x_i,v_i,C_i,F_i,\mu_i,\lambda_i)$ for each particle; grid mass and momentum are rebuilt at every substep. Its elastic stress contribution is
\begin{equation}
 P_iF_i^\top=2\mu_i(F_i-R_i)F_i^\top+\lambda_i(J_i-1)J_iI,
\end{equation}
where $R_i$ is the polar rotation of $F_i$ and $J_i=\det(F_i)$. This is the fixed-corotated expression implemented in the source, even though older metadata calls the simulator Neo-Hookean. Singular values of $F$ are clipped to $[0.2,4]$, positions to $[0.02,0.98]$, and grid boundary velocities are zeroed.

The grid has $32\times32$ cells, particle spacing is 0.026, and an observation step contains eight substeps of size $10^{-3}$. Gravity is zero and Poisson's ratio is 0.2. After grid-to-particle transfer, the actuator contribution is applied as
\begin{equation}
 v_i\leftarrow 0.92\left(v_i+\Delta t\,55\,w_i(c)\,f\right),\qquad
 w_i(c)=\frac{\exp(-\|x_i-c\|^2/(2r^2+\epsilon))}{\sum_j\exp(-\|x_j-c\|^2/(2r^2+\epsilon))+\epsilon},
\end{equation}
with $r=0.10$ and $\epsilon=10^{-8}$. These are dimensionless simulator force commands; no real-world force calibration is implied. The weight normalization makes the total applied contribution nearly independent of actuator center, while $c$ changes its spatial distribution. A finite-difference audit on 16 validation states at $t=3$ gives median one-step RMS-position sensitivities of $1.48\times10^{-6}$, $0.91\times10^{-6}$, $2.36\times10^{-6}$, and $2.36\times10^{-6}$ for perturbations of $c_x,c_y,f_x,f_y$, respectively. After scaling by the training-set standard deviation of each action coordinate, typical force-coordinate effects are about twice the center-coordinate effects. The proximity diagnostic $\min_i\|x_i-c\|\leq2r$ is true for all 3,000 validation transitions, so it is not used to claim contact-stratified robustness.

The $0.92$ multiplier is applied on each of eight substeps, giving a nominal $0.92^8=0.513$ velocity factor per observation in the absence of other updates. In a one-impulse-then-zero-action control on 16 validation states, the median RMS-velocity ratio at horizon 5 versus horizon 1 is 0.163 under the paper setting; reducing the damping parameter from 0.08 to 0.03 and 0.01 increases that ratio to 0.394 and 0.707. These controls characterize the environment and are not additional trained-model comparisons.

Rendering rounds $(63x_i,63y_i)$ to pixel centers and adds local Gaussian splats with $\sigma=1.35$ pixels. Density is divided by its framewise maximum. Velocity channels use $0.5+0.5\tanh(v)$ after local density normalization. Values are clipped to $[0,1]$, multiplied by 255, and cast to uint8. Thus spatial rounding, velocity compression, and channel quantization can each hide small counterfactual changes. Framewise density normalization also removes absolute mass scale.

\begin{figure}[ht]
\centering
\includegraphics[width=.77\linewidth]{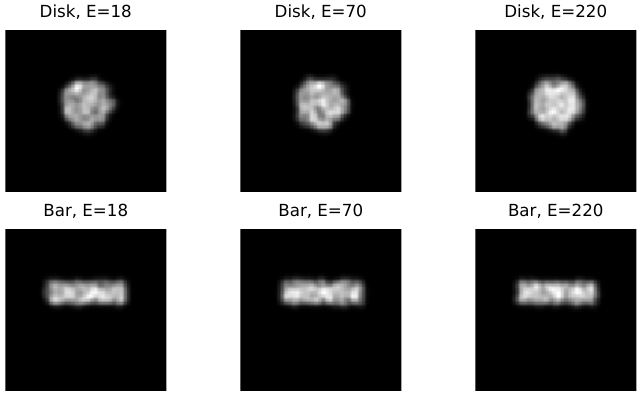}
\caption{\textbf{Density-channel examples from the recorded-action replay.} Episodes 1800--1805 at observation 12, covering both shapes and all three modulus values. These are simulator observations, not model reconstructions. The small source array is packaged so this figure regenerates without the full tensor.}
\label{fig:examples}
\end{figure}

\section{Training and reproducibility details}
\label{app:repro}
\begin{table}[ht]
\centering
\caption{Corrected training configuration shared across the reported checkpoints. The no-action model replaces action fusion with a state-only projection.}
\small
\begin{tabular}{p{.38\linewidth}p{.54\linewidth}}
\toprule
Component & Configuration \\
\midrule
Encoder & ViT; 4 blocks; width 128; 4 heads; MLP ratio 2 \\
Patches / representation & $8\times8$ patches; CLS token; latent width 128 \\
Predictor & 4 residual-MLP blocks; width 128; MLP ratio 2 \\
Action embedding & Width 64; concatenation fusion \\
EMA & Momentum 0.996; updated each training step \\
Dropout / stochastic depth & 0 / 0.05 \\
Optimizer & AdamW; $\beta_1=0.9$, $\beta_2=0.999$; weight decay 0.05 \\
Schedule & 200 epochs; 20 warmup; cosine; LR $2\times10^{-4}$ to $10^{-6}$ \\
Batch / clipping & 128 / gradient norm 1.0 \\
VICReg coefficients & Invariance 25; variance 25; covariance 1 \\
Variance hinge & Target standard deviation 1; ridge $10^{-4}$ \\
Data preprocessing & No random crop; jitter 0.4; blur probability 0.25; maximum blur $\sigma=2$ \\
Model selection & Final epoch-200 \texttt{last.pt} for every comparison \\
\bottomrule
\end{tabular}
\end{table}

The variance hinge and covariance penalty are summed over prediction and stop-gradient target tensors. Their target-branch values contribute to the logged objective but not its gradient. This differs from jointly regularizing two trainable encoders and is relevant when interpreting spectral outcomes. Photometric transformations are temporally consistent, but applied to physical channels; the study does not establish that these are physically valid augmentations.

Training occurred on an Apple Silicon MPS system with 8~GiB unified memory. The six newly completed 200-epoch jobs logged 2.60--2.73 hours each (approximately 16.0 accelerator-hours total). New checkpoint evaluation and fixed probes run on CPU with two PyTorch threads. The packaged run records contain evaluation durations and versions. Original environment records report Python 3.13.2, PyTorch 2.12.1, and NumPy 2.5.2; the replay evaluator records its Linux environment separately. Training and evaluation seeds are distinct: training uses 42--45, data initialization uses 0, and shuffling/probe initialization uses 0.

The ordinary manuscript source archive includes model and simulator code, configs, the canonical validation-only evaluator, recorded actions and labels, cached counterfactual metrics, Phase-2 control outputs, and figure/table regeneration scripts, but omits the large observation tensor and checkpoints to keep the source package compact. A separate reproducibility artifact provides the original observation tensor,
all ten final checkpoints, source code, cached metrics, manifests, and
reproduction instructions at
\url{https://osf.io/t432p/overview?view_only=c1c422ff37234321a89cee348dd381e8}. The artifact ZIP has SHA-256
\begin{center}\footnotesize\texttt{7982e886e3d5a4bd92fb31feb52a13dc885042eaf405f245d727e9774b98de04}.\end{center}
Thus exact checkpoint evaluation is available without retraining. Original project code in that artifact is released under the MIT License; the adapted MLS-MPM kernel retains the upstream MIT notice for the 88-line reference implementation associated with Hu et al.~\citep{hu2018mlsmpm}, as documented in the artifact's \texttt{THIRD\_PARTY\_LICENSES.md} and \texttt{LICENSE\_AUDIT.md}.

\section{Data provenance and evaluation scope}
\label{app:provenance}
\paragraph{Original experiment.}
The original observation array has recorded shape $(2000,16,3,64,64)$, type uint8, and SHA-256
\begin{center}\footnotesize\texttt{ee3fe0fedfc65b690c5de5526a4a434a5dc3a44ee9f9064da7317b5f4cb0759f}.\end{center}
The original simulator-fork CSVs are preserved unchanged. Their archive contains 54,600 physical rows, 109,200 target-latent rows, and 42,000 predictor rows across primary and secondary interventions. Keys are unique at their declared episode/time/intervention/horizon/checkpoint scope. There are 200 distinct validation episodes. A deterministic local replay using the original episode RNG and actions was run over all 2,000 episodes: 0 of 32,000 frames, 0 of 393,216,000 channel-pixel values, 0 episodes, and 0 actions differed from the original dump; the maximum uint8 difference was 0. This establishes local bitwise self-reproduction of the dump, not cross-platform identity.

\paragraph{Earlier Linux recorded-action replay.}
An earlier revision regenerated observations on Linux while fixing the supplied recorded actions. Its observation SHA-256 is
\begin{center}\footnotesize\texttt{aedda53c0486f2dea3903d78bf360208e57040665defbeb00064180716485a7b}.\end{center}
The original and Linux-replay action arrays share SHA-256
\begin{center}\footnotesize\texttt{cde17f18b4c123086297a59eea465ed75730f2ea7751fe7e6c1d6ef5ea70f9d1}.\end{center}
Geometry and stiffness label checksums also match. Because the Linux observation checksum differs, we do not call that replay byte-identical. It is retained only as a cross-platform consistency check; the final ten-checkpoint table and Phase-2 controls use the original dump.

\begin{table}[ht]
\centering
\caption{Cross-platform consistency checks between original cached metrics and the earlier Linux recorded-action replay. The local original-RNG replay is separately bitwise identical to the original dump.}
\label{tab:consistency}
\small
\begin{tabular}{llrr}
\toprule
Metric & Seed & Original & Replay \\
\midrule
Mean $D_{\mathrm{pred}}\times10^6$ & 42 & 3.1832 & 3.1832 \\
Mean $D_{\mathrm{pred}}\times10^6$ & 43 & 4.1197 & 4.1197 \\
Mean recorded-future error & 42 & 0.07686 & 0.07686 \\
Mean recorded-future error & 43 & 0.07025 & 0.07025 \\
Predicted effective rank & 42 & 12.268 & 12.268 \\
Predicted effective rank & 43 & 5.776 & 5.776 \\
\bottomrule
\end{tabular}
\end{table}

\paragraph{Split correction.}
The earlier evaluator computed rollout MSE over all 2,000 episodes while computing spectral metrics and probe accuracy on the 200-episode validation split. The canonical evaluator now restricts rollout error, variance normalization, and spectral measurements to episodes 1800--1999 of the original dump. It initializes forecasts from actual frame 0, uses 10 recorded controls, and evaluates target embeddings at future frames 1--10. All ten checkpoints use this identical protocol. Old pooled rollout errors and the legacy cropped objective comparison are superseded.

\paragraph{Class counts and baselines.}
Training geometry counts are 900/900 and validation counts are 101/99; stiffness counts are 600/600/600 and 67/67/66. For contact quadrants over forecast horizons 1--10, validation counts are 853/829/151/167. We choose each constant baseline from the training majority at that horizon, yielding validation accuracies $[36.5,36.5,36.5,41,41,41,42.5,42.5,42.5,41]\%$, or 40.1\% on average. Choosing the most frequent validation label after seeing validation labels would be a different procedure. Probe horizon accuracies share episodes and are not independent replications.

\section{All corrected checkpoint results}
\label{app:fullresults}
\begin{table}[ht]
\centering
\caption{Every corrected epoch-200 checkpoint on the original validation dump. VN@10 denotes target-variance-normalized error. PredR@1 and TgtR@1 are horizon-1 effective ranks for prediction and EMA-target branches.}
\small\setlength{\tabcolsep}{2.5pt}
\begin{tabular}{llrrrrrr}
\toprule
Model & Seed & MSE@1 & MSE@5 & MSE@10 & VN@10 & PredR@1 & TgtR@1 \\
\midrule
\csname @@input\endcsname tables/per_seed_rows_phase2.tex
\bottomrule
\end{tabular}
\end{table}
\begin{table}[ht]
\centering
\caption{Per-horizon effective rank for prediction (P) and EMA target (T). Pooled horizon rank is omitted here because it can inflate apparent diversity.}
\scriptsize\setlength{\tabcolsep}{2.2pt}
\begin{tabular}{llrrrrrrrr}
\toprule
Model & Seed & P1 & T1 & P3 & T3 & P5 & T5 & P10 & T10 \\
\midrule
\csname @@input\endcsname tables/rank_horizon_rows.tex
\bottomrule
\end{tabular}
\end{table}

\begin{table}[ht]
\centering
\caption{All corrected forecast probes and action interventions on the original validation dump. Accuracies are percentages averaged over horizons 1--10. The last column is the signed percentage change in one-step recorded-future error at command changes. No-action zeros are structural; exact machine values remain in the packaged CSV.}
\small\setlength{\tabcolsep}{4pt}
\begin{tabular}{llrrrr}
\toprule
Model & Seed & Geometry & Stiffness & Contact & Shuffle change (\%) \\
\midrule
\csname @@input\endcsname tables/probe_rows.tex 
\bottomrule
\end{tabular}
\end{table}

\section{Additional original simulator counterfactuals}
\label{app:secondary}
The same-time shuffle preserves the per-time marginal action distribution while changing the source episode. Permutations use NumPy Generator seed $1009t$. A single fixed point is swapped with the first available other index; multiple fixed points are cyclically shifted. The normalized mean action is not a zero-force command: it is the four-dimensional training-split raw mean. Boundary results exclude $t=0$, which has no preceding command.

\paragraph{Visibility-conditioned and state-sensitivity controls.}
The high-visibility subset is the top third of nonzero one-step raster MSE values and contains 579 transitions. We normalize target and predictor contrasts by each branch's mean validation variance. For the state-sensitivity control, a fixed isotropic random direction (seed 0) perturbs $z_t$ with norm $\sqrt{dD_{\mathrm{tgt}}}$ while the action is held fixed. Table~\ref{tab:phase2cf} shows that variance normalization strengthens the response gap and that the predictor changes far more under matched state perturbations than under action replacement.
\begin{table}[ht]
\centering
\caption{Phase-2 action-sensitivity controls. Raw and VN are median $D_{\mathrm{pred}}/D_{\mathrm{tgt}}$ on high-visibility pairs. State/action ratios are medians for $D_{\mathrm{tgt}}>10^{-8}$ and high-visibility subsets.}
\label{tab:phase2cf}
\small\setlength{\tabcolsep}{3pt}
\begin{tabular}{rrrrrrrr}
\toprule
Seed & $n_{vis}$ & $D_{tgt}$ & $D_{pred}$ & Raw & VN & State/action & State/action$_{vis}$ \\
\midrule
\csname @@input\endcsname tables/phase2_counterfactual_rows.tex
\bottomrule
\end{tabular}
\end{table}

\begin{table}[ht]
\centering
\caption{All original one-step intervention families. Position and velocity columns are medians in simulator units. The first two rows are primary contrasts. Secondary force/contact perturbations are not pooled into the primary conclusions.}
\small
\begin{tabular}{lrrrr}
\toprule
Intervention & Pairs & RMS pos. $\times10^5$ & RMS vel. $\times10^3$ & Identical raster (\%) \\
\midrule
\csname @@input\endcsname tables/secondary_rows.tex 
\bottomrule
\end{tabular}
\end{table}

Secondary interventions reverse force, rotate it $90^\circ$ counterclockwise, scale it by 1.5, or shift $c_x$ by 0.10 with clipping to $[0.08,0.92]$. The snapshot and command-hold protocol is otherwise identical. These interventions probe a different action distribution and are reported as diagnostics. The per-transition target and predictor files include all families; figure and table regeneration do not require checkpoint inference.

The response ratio uses the strict condition $D_{\mathrm{tgt},1}>10^{-8}$. Its valid sample sizes are 1,727 (seed 42) and 1,744 (seed 43), excluding 1,273 and 1,256 of 3,000 pairs. These exclusions are selected: 97.9\% of the seed-42 excluded pairs have identical rasters, whereas the included group has larger action distance and particle displacement. We therefore keep ratios secondary to unnormalized and visibility-conditioned contrasts. Conditional median raw ratios are 0.0165 and 0.0670. At horizons beyond one step, only physical and target-latent divergences were measured; no corresponding multi-step predictor response is claimed.

For original counterfactual means, 5,000 percentile bootstrap resamples draw whole validation episodes with replacement (seed 0), retaining transitions within each episode. The statistic is the mean of per-episode means. These intervals describe episode-sampling uncertainty conditional on a checkpoint, not training-seed uncertainty. The seed-42 95\% interval for mean predictor contrast is $[3.03,3.34]\times10^{-6}$. The action-distance/physical-displacement Spearman correlation is $\rho=0.684$ with a 95\% episode-bootstrap interval $[0.659,0.707]$. Original all-transition changes in recorded-future error are $+0.015\%$ and $+0.013\%$ for seeds 42/43. Two-seed objective figures show the individual runs directly rather than treating their range as a confidence interval.

\section{Exploratory stiffness and geometry controls}
\label{app:probes}
\begin{figure}[ht]
\centering
\includegraphics[width=\linewidth]{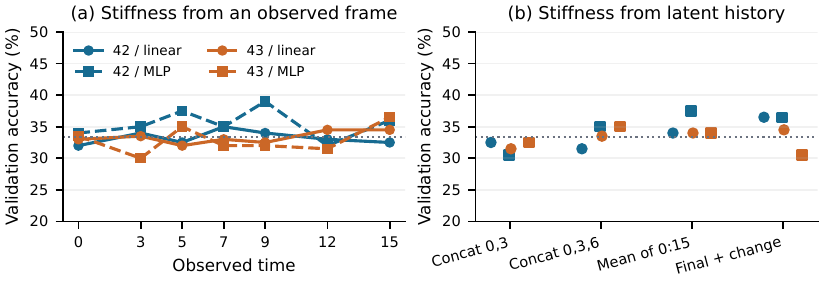}
\caption{\textbf{Fixed readouts of observed EMA embeddings on the replay.} Individual conditions for corrected VICReg seeds 42/43. Dashed horizontal lines show $1/3$ uniform stiffness chance. Lines connect descriptive accuracies, not independent samples or temporal learning curves. No validation-based choice of a best time, probe, or history is used for a primary claim.}
\label{fig:probecontrols}
\end{figure}

Both linear and MLP classifiers use seed 0, AdamW with learning rate $10^{-2}$ and weight decay $10^{-4}$, and 80 full-batch updates. The MLP is $d\to128\to C$ with ReLU. Single-frame times are $\{0,3,5,7,9,12,15\}$. Latent histories concatenate times 0/3 or 0/3/6, average all 16 embeddings, or concatenate $z_{15}$ with $z_{15}-z_0$. Raw frames are area-downsampled to $16\times16$; raw history concatenates times $\{0,3,6,9,12,15\}$. These analyses were executed during the revision using the fixed repository probe specification. They were not preregistered confirmatory tests.

\begin{table}[ht]
\centering
\caption{All probe-condition ranges. ``Conditions'' counts settings per task, including both model seeds where applicable. These ranges summarize the full packaged 120-row table, not error bars or confidence intervals.}
\small
\begin{tabular}{llrrr}
\toprule
Input & Readout & Conditions & Stiffness (\%) & Geometry (\%) \\
\midrule
\csname @@input\endcsname tables/observability_rows.tex 
\bottomrule
\end{tabular}
\end{table}

The earlier $16\times16$ raw-history MLP's 50.5\% geometry accuracy is a failed positive control, so that particular negative stiffness result remains inconclusive. The new upstream hierarchy in Table~\ref{tab:stiffnesspositive} resolves the broader interpretation: full-resolution float and uint8 rasters remain essentially at the stiffness majority baseline while geometry is perfectly linear-decodable; mechanical $(x,v,C,F)$ readouts are unstable and never consistently separate the three stiffness classes; privileged material parameters yield 100\% stiffness with the MLP. The supported conclusion is therefore weak material identifiability under the tested excitation and observation regime, not encoder-specific stiffness loss. Carefully designed material interventions and longer excitation remain necessary for a stronger material-grounding claim.

\begin{table}[ht]
\centering
\caption{Upstream positive controls for material identifiability. Ranges span fixed linear/MLP readouts and tested times/histories. The validation-majority stiffness baseline is 33.5\%. Privileged $(\mu,\lambda)$ contains the defining material parameters and is an oracle control.}
\label{tab:stiffnesspositive}
\small
\begin{tabular}{lrr}
\toprule
Input & Stiffness (\%) & Geometry (\%) \\
\midrule
\csname @@input\endcsname tables/stiffness_positive_rows.tex
\bottomrule
\end{tabular}
\end{table}

\section{Claim boundaries and broader impact}
The study reports a diagnostic failure pattern within one implemented model family and simulator. It does not establish state-of-the-art physical simulation, intrinsic limitations of JEPA, causal identification of material parameters, or useful planning. In particular, a magnitude ratio does not verify response direction, and weak stiffness decoding does not prove absence of physical signal. The intended benefit is more rigorous evaluation of world-model claims; the main risk is overgeneralizing a small synthetic experiment to real physical intelligence. Explicit metric scope and preserved numerical provenance help limit that interpretation. The work uses synthetic data and no human-subject or private information.